# Reasoning about Affordances:
# Causal and Compositional Reasoning in LLMs

**Magnus F. Gjerde\*, Vanessa Cheung\*, & David Lagnado\***
\*Department of Experimental Psychology, University College London
ucjumfg@ucl.ac.uk

## Abstract

With the rapid progress of Large Language Models (LLMs), it becomes increasingly important to understand their abilities and limitations. In two experiments, we investigate the causal and compositional reasoning abilities of LLMs and humans in the domain of object affordances, an area traditionally linked to embodied cognition. The tasks – designed from scratch to avoid data contamination – require decision-makers to select unconventional objects to replace a typical tool for a particular purpose, such as using a table tennis racket to dig a hole. In Experiment 1, we evaluated GPT-3.5 and GPT-4o, finding that GPT-4o, when given chain-of-thought prompting, performed on par with human participants, while GPT-3.5 lagged significantly. In Experiment 2, we introduced two new conditions, Distractor (more object choices, increasing difficulty) and Image (object options presented visually), and evaluated Claude 3 Sonnet and Claude 3.5 Sonnet in addition to the GPT models. The Distractor condition significantly impaired performance across humans and models, although GPT-4o and Claude 3.5 still performed well above chance. Surprisingly, the Image condition had little impact on humans or GPT-4o, but significantly lowered Claude 3.5's accuracy. Qualitative analysis showed that GPT-4o and Claude 3.5 have a stronger ability than their predecessors to identify and flexibly apply causally relevant object properties. The improvement from GPT-3.5 and Claude 3 to GPT-4o and Claude 3.5 suggests that models are increasingly capable of causal and compositional reasoning in some domains, although further mechanistic research is necessary to understand how LLMs reason.

## 1  Introduction

For a long time scholars have debated whether our conceptual system is grounded in embodiment and experience (Lakoff & Johnson, 1982; Harnad, 1990). The remarkable progress of large language models (LLMs; Zheng et al., 2024; Lu et al., 2024; Zhong et al., 2023) in a variety of tasks, despite these models being trained on text and static images without direct, embodied experience in the real world, has rekindled this debate (Harnad, 2024; Pavlick, 2023). This debate is unlikely to be resolved soon, but an intriguing way to explore the issues it raises is to test LLMs in domains that are closely connected to embodiment and interactive experience with the real world. Are LLMs able to perform well on tasks that have been thought to depend on embodied experience (Glenberg & Robertson, 2000)? One such approach is to test LLMs in the domain of object affordances and their capacity to reason about objects and their functional properties.

Recent studies suggest that LLMs have developed increasing sensitivity to object affordances (Jones et al., 2022; Yiu et al., 2023), which may suggest that they are capable of solving such tasks without embodiment. However, as LLMs are trained on vast datasets containing extensive knowledge about objects and their uses, they might simply be retrieving memorised patterns in the training data when solving object affordance tasks. This ties in with a central debate about LLMs' reasoning abilities: do they reason in robust ways that generalise, or do they mainly memorise (Geirhos et al., 2020; Krakauer & Mitchell, 2023; McCoy et al., 2023)? To test LLMs' ability to reason and generalise about object affordances, not just memorise typical object uses, we need novel tasks that prevent LLMs from relying on training data patterns.

In the present study, we address this through a tool innovation task based on Yiu et al. (2023). In our experiments, participants (humans and LLMs) must select alternative objects to accomplish goals typically achieved using specific tools (e.g., hammering a nail without a hammer). To make the task more difficult, the alternative options include objects associated with the typical tool (e.g., a screwdriver). Crucially, the afforded (correct) object is not related to the typical tool (a saucepan, in this case), forcing the LLMs to disregard co-occurrences and associations in the training data and make a choice based on causally appropriate properties (e.g., weight, solidity, and flatness). We present this task to LLMs across different modalities and levels of difficulty. We suggest that success requires both causal reasoning (Sloman & Lagnado, 2015) about which properties are causally relevant and compositional reasoning - the ability to decompose concepts into abstract elements and recombine them (Lake et al., 2017).

**Figure 1**
*Illustration of the Compositional Structure of a Tool Innovation Task*

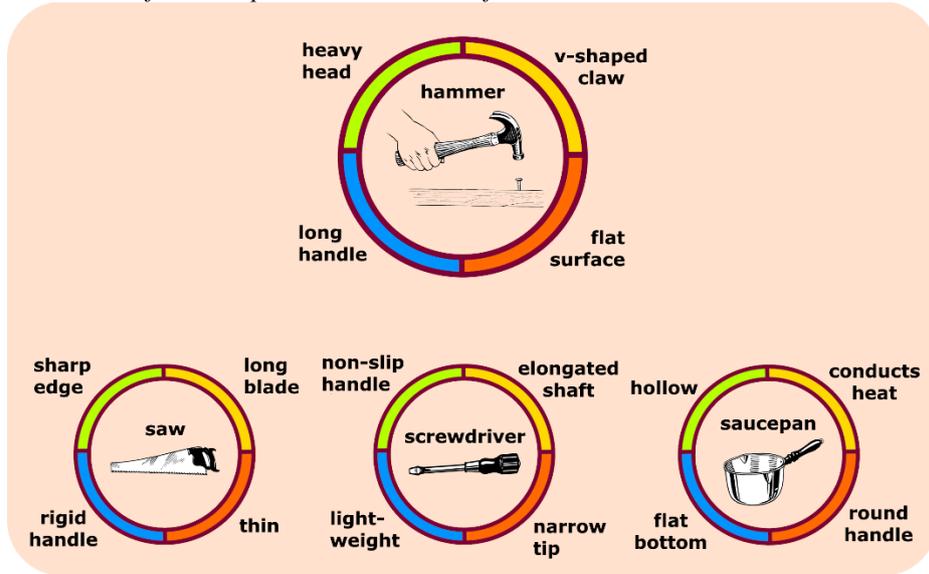

*Note*. In this example, the goal is to replace a hammer with one of the objects below. To find the correct object, the participant must be able to decompose the objects into their constituent abstract properties, such as flatness, hardness, and shape, and determine which properties are superfluous and which are causally necessary for the task.

## 2 Background and Related Work

**Causal reasoning and generalisation** There have been several studies examining the causal reasoning abilities of LLMs, some positive (Lampinen et al., 2023; Thagard, 2024) and some negative (Joshi et al., 2024; Bao et al., 2024; Binz & Schulz, 2023). Kıcıman et al. (2023) examined multiple varieties of causal reasoning in LLMs, including causal discovery, effect inference, and causal attribution. Results showed that fairly recent LLMs, such as GPT-4, perform remarkably well across the tasks compared to earlier LLMs, even approaching human baseline performance in one test of counterfactual reasoning. Jin et al. (2023) examined LLMs' formal causal reasoning based on Pearl's causal framework (Pearl & Mackenzie, 2018). GPT-4, the best performing model, achieved an accuracy of 64.28%, roughly 14% above chance. On the interventional tasks, GPT-4 scored 81.87%, suggesting that causal reasoning can, to some degree, be learnt passively (see also Lampinen et al., 2023).

However, as Kıcıman et al. (2023) emphasise, it is unclear how much of the LLMs' accuracy reflects general reasoning and how much is a result of retrieval of memorised patterns. Larger models tend to perform better, and this may be due to them generalising better than smaller models. Generalisation concerns a cognitive system's ability to deal with novelty. It is not a discrete either-or ability, but a spectrum (Chollet, 2019, p. 9-12). Testing LLMs' ability to generalise is typically done in one of three main ways: by introducing novel examples of tasks with familiar structures (e.g., Zhang et al., 2024), by designing tasks with entirely new structures (e.g., Wu et al., 2023), or by creating tasks that challenge LLMs to navigate a tension between associative retrieval and rule-based reasoning (e.g., Yiu et al., 2023). Associative retrieval involves recognizing and recalling previously encountered patterns, whereas rule-based reasoning requires understanding broad invariant principles.

Wu et al. (2024), for example, created a dual test set composed of standard and structurally modified tasks. Results showed that GPT-4, GPT-3.5, Claude, and PaLM all performed significantly worse in the non-standard tasks compared to the standard tasks (Wu et al., 2024). However, the performance decrease is substantially less steep for GPT-4 than for the other models. This suggests that the state-of-the-art model (at that time) may be capable of some general reasoning. Other studies have found a similar pattern, where the latest models are more capable of solving novel tasks than smaller models (for an example in analogical reasoning, see Webb et al., 2023; 2024, for one in maths, see Zhang et al., 2024).

**Affordances and tool innovation** Sensitivity to affordances (i.e., action-possibilities available to an agent in an environment; Gibson, 2014) is closely connected to causal knowledge, as knowing what is and is not afforded means being sensitive to objects' causal properties. To know what can be done with various kitchen utensils requires knowledge about the effects of sharpness, shape, solidity, weight, and so on.

In a study comparing humans and a statistical natural language processing (NLP) model, Glenberg and Robertson (2000) showed that humans were sensitive to the difference between afforded and non-afforded sentences, while the NLP model, Latent Semantic Analysis, was not. The study involved asking participants and the NLP model to judge sentences such as "He used his shirt/glasses/towel to dry his feet." More recently, Jones et al. (2022) tested language models on Glenberg and Robertson's tasks. They found that GPT-3 was sensitive – although not as sensitive as humans – to the difference between afforded and non-afforded words. While GPT-3 was trained exclusively on text, it was still to some degree able to identify that drying your feet with glasses is more surprising than doing it with a shirt. The two other models tested, BERT and RoBERTa, were not sensitive to this difference.

In a study which forms the inspiration for our experiments, Yiu et al. (2023) examined LLMs' ability to solve a tool innovation task. In their experiment, humans and LLMs were presented with a goal typically achieved with a specific tool (e.g., drawing a circle without a compass) and three alternative object options: 1) a superficially similar but functionally inappropriate object (ruler), 2) a superficially different but functionally suitable object (round-bottomed teapot), and 3) an irrelevant object (stove). In the innovation task, participants had to select the alternative object that would best accomplish the given goal. In the imitation task, they were asked to select which object would "go best" (i.e., which object is most associated) with the typical tool.

Unlike selecting an associated object, Yiu et al. (2023) argued that discovering new applications for everyday objects requires going beyond statistical co-occurrence patterns ("imitating"); it is to appreciate "the more abstract functional analogies and causal relationships between objects that do not necessarily belong to the same category or are associated in text" (2023, p. 5; see also Jones et al., 2022).

Their results showed that humans and LLMs performed well on the imitation task (84.9% for humans vs. 83.3% for GPT-4). However, in the innovation task, human children (85.2%) and adults (95.7%) scored higher than GPT-4 (75.9%) and GPT-3.5-turbo (58.9%). Yiu et al. write that these results suggest that LLMs' text-based learning "may not be sufficient to achieve tool innovation" (p. 5). However, the distance between the performance of children and adults (10.5%) is greater than the distance between children and GPT-4 (9.3%), but the authors do not argue that children struggle with tool innovation. Rather than seeing the results as evidence that LLMs cannot innovate, we suggest that the results show significant progress since Jones et al. (2022) identified a moderate affordance sensitivity in GPT-3.

**Capability-oriented evaluation** One challenge with evaluating both natural and artificial cognitive systems is that test performance may be informative, but it is not always clear in what sense it is informative. Did the system use a robust and generalisable strategy to solve the tasks, or did it use a shortcut that fails on task variations (Geirhos et al., 2020)? It is important to distinguish between a system's performance on a given number of tasks and the system's underlying capability that enables the system to solve that type of task generally (Rutar et al., 2024; Burden et al., 2023).

This distinction highlights the need to be explicit about the underlying capabilities that a benchmark or evaluation is designed to measure. Yiu et al. (2023) argued that solving the tool innovation tasks required sensitivity to abstract functional analogies, causal relations between objects, and the ability to make out-of-distribution generalisation. This is highly plausible, but it raises the question of how the out-of-distribution generalisation is made.

We argue that this is done through what Lake et al. (2017) call compositional reasoning: the ability to break concepts into abstract parts and subcomponents and to recombine these into new configurations. To determine which object can replace a typical tool, one must be able to decompose the typical tool into a set of abstract properties that can be found in other objects. Causal knowledge is necessary to know which of the multiple properties are necessary for the task and which are superfluous. Figure 1 shows a simplified example of the potential object properties in such a task.

Thus, in this article, we view the tool innovation task through the lens of causal and compositional reasoning. LLMs' limitations in rule-bound compositional reasoning (Dziri et al., 2023) do not preclude a softer form of approximate compositional reasoning that involves feature decomposition and recombination (Lepori et al., 2023).

## 3 Present studies

Our two experiments investigate LLMs' causal and compositional reasoning abilities in the domain of object affordances. In both experiments, we present human participants and LLMs with a task comprised of 20 tool innovation questions. In each task, the decision-maker must accomplish a given goal in the absence of a typical tool used to accomplish this goal. The question is whether the decision-maker can select an appropriate object to accomplish the goal, even though it is not typically used for that purpose. In our revision of Yiu et al.'s (2023) paradigm, we increase the difficulty of the task by including another associated but inappropriate object, for a total of four options. This makes random answers less likely to succeed.

In Experiment 1, we test OpenAI's GPT-3.5-turbo and GPT-4o models. We vary their temperature settings (which determines how variable or deterministic the model's output is), allowing us to examine the robustness of the models' reasoning skills. We also test the models with a chain-of-thought (CoT) prompt, as CoT prompting has been shown to improve LLM performance in several domains (Kojima et al., 2022; Wei et al., 2022). In Experiment 2, we test two further LLMs in addition to the GPT models, namely Anthropic's Claude 3 Sonnet and Claude 3.5 Sonnet models, and we add two new experimental conditions: Distractor and Image. The Distractor condition adds five further object options to the existing four options, making the task harder. The Image condition uses images instead of words to present the four object options. We also examine the effect of CoT prompting in Experiment 2.

Finally, one limitation of past research is that it tends to focus on reporting quantitative performance scores, without going into qualitative detail about how LLMs responded to queries. This makes it difficult to evaluate what different AI models got right and wrong. Therefore, in both experiments, we conduct qualitative analyses of the answers generated by the models to better characterise their performance differences.

## 4 Experiment 1

### 4.1 Method

**Participants** This study was preregistered on AsPredicted.org (#184165). This study received ethical approval from the UCL Psychology Ethics Committee under code EP/2018/005. We recruited 100 participants from the United Kingdom using the online platform Prolific. The age range was 19-81 years (M = 42.3, SD = 12.7), with 50 male and 50 female participants. We paid participants £8.62/h for the 6-minute study. All participants passed the attention checks, resulting in no exclusions.

**Materials** The experiment consisted of 20 tool innovation tasks presented to both human participants and LLMs (see Appendix A). Each task required achieving a goal without a typical tool used for that purpose. Participants chose from four options: two objects associated with but unsuitable for the task ("associated objects"), one unrelated but suitable object ("afforded object"), and one irrelevant object ("distractor object"). Associated objects were chosen on the basis of use in similar contexts (e.g., a plate is associated with a glass). An example of a task is presented below:

**Figure 2**
*An Example Question in the Tool Affordance Task*

| |
|---|
| Your task is to take a baked and hot cake out of the oven. Normally, you would use oven mittens to accomplish this task. However, oven mittens are not available to you. At your disposal, you have the objects listed below. Which of these would you use to accomplish the task? |
| A balloon (irrelevant)      A saucepan lid (associated) |
| A beach towel (afforded)    A chef's hat (associated) |

**Design** We presented the same task to human participants and the LLMs. The main analysis was a between-subjects comparison of the response accuracy of three groups (humans, GPT-3.5, and GPT-4o). All participants and models answered the same set of questions (i.e., questions were shown within subjects for participants; for more details, see Data Analysis section below). For the LLMs, responses were collected at three temperature settings (0, 1, and 2). The standard prompt was presented to all temperature variants, but the CoT prompt was only presented to the models at the standard temperature 1. We collected 25 responses per model at each of the different temperatures and 10 responses per model using the CoT prompt.

**Procedure** We presented the task items in random order to the human participants. Participants saw two attention checks in the experiment, which were simple multiple-choice questions about the common function of certain tools. To gain insight into the problem-solving strategies employed by humans, we included at the end of the survey a multiple-choice question about how they approached solving the tasks. LLMs were presented with the tasks independently, in other words, not within the same context window. The OpenAI GPT models were accessed via API using the openai package in Python. The prompts to the GPT models contained the same text seen in the example above, also with randomization of option order, but with the following addition at the end: "Only specify the chosen object: ". This resulted in simple answers containing only the chosen object. For the chain-of-thought prompt, this last sentence was modified to the following: "Evaluate the suitability of each object separately. Then, specify your choice." This prompt resulted in long answers.

**Data analysis** The LLM scores were averaged by model run, which means that the model's responses to the 20 questions were averaged for each round, and the resulting averages were treated as individual data points. This approach, inspired by Cheung et al. (2024), was adopted to enable statistical comparison with human responses. For the overall comparison of the accuracy means, GPT-3.5 and GPT-4o's scores across temperatures and prompts were aggregated.

The aggregation of LLM scores by model run has both advantages and shortcomings. It provides a practical way to statistically compare human and LLM performance. However, unlike human responses, which are independent samples, LLM responses are stochastic outputs from the same model. This aggregation method treats each model run as if it were an independent participant, which is not strictly accurate, but serves our analytical purposes. Doing so also helps highlight the variance in LLM responses, which would be hidden in a single average accuracy metric. For LLM accuracy on a per-question basis, all their answers were averaged irrespective of model run. For all statistical tests, we applied the Bonferroni correction.

### 4.2 Results

**Accuracy comparisons** The mean accuracy of humans and the GPTs with different temperatures and prompts is shown in Figure 3 below. To test whether there were significant differences between the aggregated means of GPT-3.5, GPT-4o, and humans, we performed a one-way ANOVA with Model Type (human, GPT-3.5-turbo, GPT-4o) as the independent variable and accuracy as the dependent variable. It showed a statistically significant effect of Model Type on accuracy, $F(2, 267) = 668.3$, $p < .001$. Follow-up pairwise comparisons, shown in Table 1, revealed significant differences between all three pairs ($p < .001$). Both models demonstrated significantly lower accuracy compared to humans, with GPT-3.5 showing the largest difference.

**Figure 3**
*Mean Accuracy of Humans and Configurations of the GPT models*

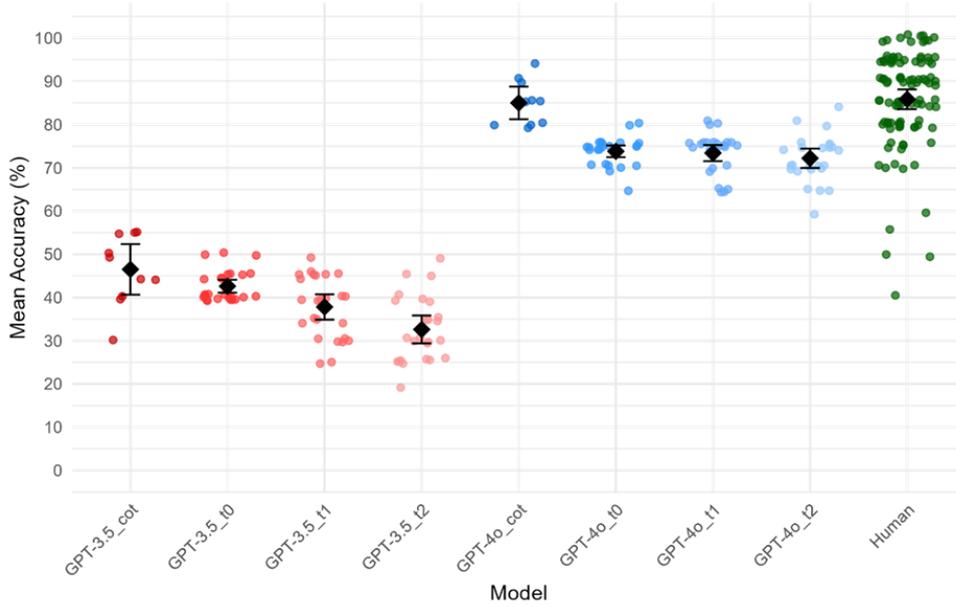

*Note.* On the x-axis, the t0, t1, and t2 labels represent the different temperature settings, while the "cot" label represents performance with chain-of-thought prompting. Black diamonds represent mean accuracy for each model, with error bars showing the 95% confidence intervals. Individual data points are jittered for better visibility.

**Table 1**
*Pairwise Comparisons of Mean Accuracy Between Humans, GPT-3.5, and GPT-4o*

| Contrast | Estimate | SE | df | t | p |
|---|---|---|---|---|---|
| GPT-4o - Human | -0.11 | 0.01 | 267 | -8.52 | < .001 |
| GPT-3.5 - Human | -0.47 | 0.01 | 267 | -35.50 | < .001 |
| GPT-3.5 - GPT-4o | -0.36 | 0.01 | 267 | -25.94 | < .001 |

*Note.* Estimate, i.e. the difference between the means, is expressed as a ratio of correct responses.

When comparing the accuracy of humans with various configurations of GPT-4o, we found significant differences between humans and GPT-4o_t0, GPT-4o_t1, and GPT-4o_t2 (all $p < .001$; see Table 2). However, there was no significant difference between human performance and GPT-4o with CoT prompting ($p = 1.00$). Indeed, the mean accuracy of GPT-4o_cot was 85%, only 0.8% lower than the mean accuracy of human responses. This shows that chain-of-thought prompting enables GPT-4o to perform significantly better and reach a similar performance level as humans on this task. As can be seen in Figure 3, GPT-4o_cot also performed significantly better than all temperature variations of GPT-4o with default prompting.

**Table 2**
*Pairwise Comparisons of Mean Accuracy Between Humans and Configurations of GPT-4o*

| Contrast | Estimate | SE | df | t | p |
|---|---|---|---|---|---|
| GPT-4o_cot - Human | -0.01 | 0.03 | 180 | -0.28 | 1.00 |
| GPT-4o_t0 - Human | -0.12 | 0.02 | 180 | -5.93 | < .001 |
| GPT-4o_t1 - Human | -0.12 | 0.02 | 180 | -6.13 | < .001 |
| GPT-4o_t2 - Human | -0.14 | 0.02 | 180 | -6.72 | < .001 |

**Effect of prompt and temperature** As CoT prompts were given to models at temperature 1, we statistically compared their CoT performance to that with a normal prompt at temperature 1. We found that CoT prompting had a significant effect for both GPT-3.5 and GPT-4o ($t(33) = 3.13$, $p = .04$ and $t(33) = 6.57$, $p < .001$). The effects of the different temperature settings are visible in Figure 3. While GPT-3.5's performance declines in a stepwise fashion from temperature 0 to 2, GPT-4o's performance remains stable, with only a small visible decrease in average performance at temperature 2.

**Human problem-solving strategies** When asked how they approached the tasks, 65% of human participants reported using visualization, 23% reported reasoning about object properties, and 12% relied on intuition. Two additional available options—"I don't know or remember how I solved the tasks" and "Other" (with space for specification)—were not selected by any participants. This suggests that the three categories selected by participants captured their perceived problem-solving strategies.

**Qualitative analysis of GPT responses** Here, we explore how GPT-3.5 and GPT-4o's responses to the chain-of-thought prompts differ. Ten responses per model were generated for each of the 20 questions. Figure 4 shows the models' accuracy on each question (see Appendix A for an overview of all the 20 questions). GPT-4o generated an average of 261 words per response, while GPT-3.5-turbo generated 163 words on average. Due to this dataset containing almost 85,000 words, we only present and analyse cases where responses differ in notable ways.

**Figure 4**
*Mean Accuracy of GPT-3.5 and GPT-4o with CoT Prompting Across the 20 Tasks*

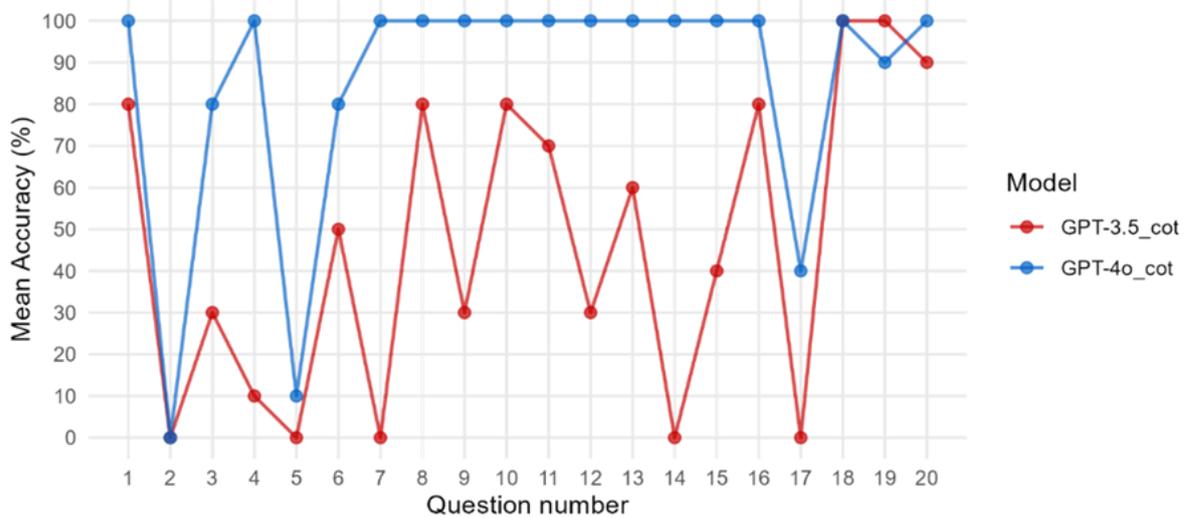

*Question 3 – hammering a nail* This task required replacing a hammer for hammering a nail. The afforded object was a saucepan, while the two associated objects were a screwdriver and a saw. GPT-3.5 scored 30% on this task and GPT-4o scored 80%. Below are examples from GPT-3.5 and GPT-4o's evaluations of the saucepan and the screwdriver for the task:

| GPT-3.5 | GPT-4o |
|---|---|
| "A **saucepan** is typically used for cooking and does not have a flat, heavy end like a hammer that would be required to effectively drive a nail into a surface." | "**Saucepans**, particularly ones with a heavy, flat base, can be used to apply a significant amount of force. The handle provides a good grip, and the flat, sturdy base can transmit force well to the nail, making it a viable substitute for a hammer." |
| "A **saucepan** is a metal object with a flat, sturdy base that could potentially be used to hammer a nail." | "The bottom of the **saucepan** is flat and broad, capable of providing a solid impact similar to that of a hammer." |
| "[A **saucepan** is] not suitable for hammering a nail as it is too flat and wide…" | "A **screwdriver** is not designed to deliver the necessary force through repeated impact. You might be able to use the handle to tap the nail lightly, but it would be inefficient ..." |
| "A **screwdriver** could potentially be used as a makeshift hammer, as it has a solid handle and a flat end that could be used to strike the nail." | **Screwdriver:** "While it has a solid, small surface, it is not designed for providing the broad, forceful impact needed to hammer a nail." |
| "While a **screwdriver** is a hard, durable object, it is not designed for hammering and may cause damage to the screwdriver or the nail." | |

Overall, GPT-3.5 demonstrated some sensitivity to the task requirements and object properties, but often lacked precision and occasionally provided contradictory responses. It did not consistently consider the importance of a flat, sturdy surface. GPT-4o, on the other hand, was precise and consistently emphasised the correct causal properties required for the task.

*Question 4 – transporting water* This task was about replacing a glass for transporting water. The afforded object was a swimming cap and the two associated objects were a plate and a fork. GPT-3.5 scored 10% on this task and GPT-4o scored 100%. Below are examples of their evaluations of the swimming cap and the plate:

| GPT-3.5 | GPT-4o |
|---|---|
| "A **swimming cap** is not suitable for transporting water as it is designed to keep water out rather than hold it in." | "A **swimming cap** is made of a waterproof material … It is flexible and can form a small pouch-like shape when held properly. It can contain and transport water effectively despite not being a rigid container." |
| "A **swimming cap** is made of waterproof material and can hold water. However, it is not the most practical choice for transporting water as it is typically worn on the head and may not be easy to handle." | **Swimming cap:** "It can hold a reasonable amount of water if you invert it to form a makeshift container." |
| **Swimming cap:** "… it is designed to be worn on the head and does not have the ability to hold water." | "A **plate** … is not very suitable for transporting water. While it can hold some amount of water if it has slightly raised edges, the water would spill easily during transportation as plates are shallow and not designed to contain liquids efficiently." |
| **Plate**: "While it may be able to hold some water, it is not an ideal vessel for transporting water as it is not designed for that purpose and may spill easily." | "If the **plate** has high sides, it might be slightly better, but overall, it is not very practical for transporting water over any distance." |
| **Plate**: "While it may not be as efficient as a glass, a plate can still hold a decent amount of water and can be used to transport it from one place to another." | |

Here, GPT-3.5 consistently noted that the plate is not a good option due to the risk of spilling water. However, it did not appreciate that the property which enables swimming caps to keep water out is the same property that can, if the cap is held upside-down, keep water inside. In other words, GPT-3.5 recognized this relevant property, but was not able to flexibly apply it to a new use case. GPT-4o, on the other hand, consistently considered and correctly applied the relevant causal properties of the objects.

*Question 5 – cutting a cucumber into cubes* Question 5 represents an interesting case, where both GPT-3.5 and GPT-4o did very poorly, but humans scored 80%. The task was to replace a kitchen knife for cutting a cucumber into cubes. The afforded object was a ruler and the two associated objects were a peeler and a whisk. Both GPT models argued that a ruler is not sharp enough, and instead they tended to opt for the peeler. They recognized that the peeler is not ideal, but still consistently viewed it as the best option. This task is arguably difficult for the following reason: it requires a fine-grained understanding that while rulers are not sharp enough to cut in general, they may be thin and hard enough to cut certain soft foods into cubes. This understanding may be difficult to acquire without embodied interaction with such objects.

*Question 7 – sweeping the floor* This task involved replacing a broom for sweeping the floor. The afforded object was a wig and the two associated objects were laundry detergent and a bucket. GPT-3.5 scored 0% on this task and GPT-4o scored 100%. Below are examples of their evaluations of the wig and the bucket:

| GPT-3.5 | GPT-4o |
|---|---|
| "A **wig** would not be useful for sweeping the floor as it is a hair accessory and not a cleaning tool." | "A **wig** is made of hair or synthetic fibers. While unconventional, the strands of a wig could potentially be used to sweep light dust or dirt if there are no other options. It could act somewhat like the bristles of a broom if used carefully." |
| **Wig:** "… it is a hairpiece and has no function in cleaning. It would not be effective in gathering dirt and would likely make a mess if used for this purpose." | "While not ideal, a **wig** with long hair could be used to gather dust and dirt more effectively than the other items available." |
| **Bucket:** "Suitable for sweeping the floor if used creatively. You could turn the bucket upside down and use it as a makeshift tool to push dirt and debris towards a corner or into a dustpan." | "A **bucket** is rigid and designed to hold things, not sweep them. It cannot be used to move debris effectively." |
| "A **bucket** could potentially be used to help collect dust and debris, but it would not be very effective in actually sweeping the floor." | |

Unlike GPT-4o, GPT-3.5 was completely unable to appreciate the functional analogy between hair and a broom's bristles. Instead, GPT-3.5 emphasised that a wig is not a cleaning tool. GPT-4o solved the task well and successfully identified the causally relevant property of the wig.

## 4.3 Discussion

As expected, there was a significant difference between the performance of GPT-3.5 and humans. Regarding the difference between GPT-4o and humans, the evidence was mixed. With normal prompting, GPT-4o performed significantly below humans. This aligns with the findings of Yiu et al. (2023). With chain-of-thought prompting, on the other hand, which Yiu et al. did not test, GPT-4o achieved a mean accuracy nearly identical to that of humans. Moreover, the models' results at different temperatures revealed that GPT-4o's reasoning was more robust and less likely to degrade when the temperature setting was increased. The large performance differences between GPT-3.5 and GPT-4o are consistent with the existing literature, which has found that larger models tend to outperform smaller, earlier models (Zheng et al., 2023; Chang & Bergen, 2023).

Finally, the qualitative review of GPT-3.5 and GPT-4o's text output revealed some interesting patterns. Both models generally identified the causally relevant properties required to solve the tasks. GPT-3.5, however, was often unable to recognize and apply these properties in the unrelated but afforded objects. Rather than put the afforded object to a new and unusual use, it tended to focus on its usual function and area of use. Success required disregarding common use cases and flexibly applying familiar properties in new ways; in other words, to generalise beyond the familiar patterns. Moreover, GPT-3.5's factual statements were at times incorrect and contradictory. In comparison, GPT-4o showed a more precise appreciation of the causally relevant object properties and a consistent ability to flexibly apply the afforded objects appropriately. It was also evident in GPT-4o's responses that its factual and causal knowledge was richer, more detailed, and less prone to error.

This finding aligns with previous research that has found that GPT-4 performs better than earlier models on tasks that are designed to prevent memorisation (Zhang et al., 2024; Wu et al., 2023). Unlike previous research, however, this study has offered a qualitative evaluation of the ways in which GPT-4o generalises better than the earlier GPT-3.5.

When asked about their problem-solving strategy, the majority of human participants reported using mental simulation, with the rest reporting reasoning about properties or using intuition. This aligns with existing scholarship on the use of visualization and simulation in human causal reasoning (Gerstenberg, 2024; Johnson-Laird, 2010; Lagnado, 2021; Sloman & Lagnado, 2015). At the same time, it raises interesting new research questions. If humans report relying mostly but not exclusively on visualisation or simulation, is there any variability in how different LLMs solve these tasks? Future behavioural research should aim to develop experiments that can help us dissociate different cognitive problem-solving strategies in humans and LLMs. One benchmark, where solving tasks often requires visualising scenes, finds that LLMs still perform well below humans (Philip & Hemang, 2024), suggesting that LLMs struggle with visualisation (see also Rahmanzadehgervi et al., 2024).

In Experiment 2, we expand this study by comparing a broader set of LLMs and by adding two further experimental conditions.

## 5 Experiment 2

The second experiment was designed as a replication and expansion of the first experiment. It had the same structure as Experiment 1, but with some modifications. It was preregistered on AsPredicted.org (#188326).

**Image and Distractor conditions** In addition to the original text-based survey with four object options per task (hereafter called the *Standard* condition), we introduced two new conditions. First, a *Distractor* condition where the decision-maker was faced with nine options in total. This consisted of four associated but incorrect options, four irrelevant options, and one afforded option (see Appendix A for an overview). Second, we added an *Image* condition where the four original object options were shown as images instead of text (see Figure 5). This allowed us to test (1) whether using images instead of text to show the options affects the accuracy of humans and LLMs, and (2) whether having more distractor options would affect the accuracy of humans and LLMs.

Our hypothesis was that the image format would not negatively impact human performance, but that it would negatively impact the accuracy of LLMs. We believed that the LLMs, being mainly text-based, would perform less well in non-text formats. If a model has similar accuracy in the standard and image conditions, this would suggest that (1) the model is able to translate the image input (in base64 format) into a text format which it then uses to solve the tasks, or (2) the model reasons in an amodal way, whereby the relevant computations are performed in a more abstract format, that may take either text or images as input.

**Figure 5**
*Example of Image Condition Question in the Qualtrics Survey*

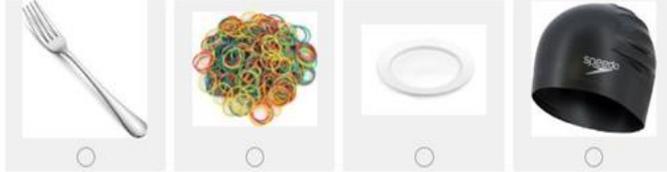

If models or humans perform significantly worse in the Distractor condition compared to the Standard condition, this could suggest that the fewer options in the Standard condition helped models and humans achieve a higher score by guessing (pure guesswork would give a score of 25%). It could also suggest that increasing the number of options is a good way of increasing task difficulty, leading to worse but not random performance. If humans or a model performs similarly in the two conditions, this would suggest that their solution is more robust and general. We hypothesized that having more options would moderately affect the performance of humans and state-of-the-art LLMs, but that it would make the earlier LLMs perform significantly worse.

## 5.1 Method

**Participants and new LLMs** For the new experiment, we recruited a representative sample of 300 participants from the United Kingdom using Prolific, which is based on UK 2021/2022 census data and stratified across age, sex, ethnicity, and political affiliation.[1] The age range of participants was 18-87 years ($M = 47.0$, $SD = 16.2$), with 141 male and 155 female participants. We paid participants £8.75/hr for the 7-minute study. We excluded four participants: one participant because their data did not come through due to technical errors, two participants who failed the attention checks, and one participant for spending less than one second answering the majority of the tasks. This left us with a sample of 296 participants, with 101 in the Standard condition, 98 in the Image condition, and 97 in the Distractor condition. In addition to testing OpenAI's GPT-3.5 and GPT-4o, we also tested Anthropic's Claude 3 Sonnet and Claude 3.5 Sonnet (hereafter referred to as Claude 3 and Claude 3.5).

**Prompts** In Experiment 1, we used a standard prompt that asked the LLMs to "only specify the chosen object," and a CoT prompt. The standard prompt restricted models to only generate the chosen option and not to elaborate on their reasoning, unlike the CoT prompt. This left open the possibility that the positive effect of CoT prompting found in Experiment 1 was to some degree due to the fact that the CoT prompt enabled longer answers, rather than the structure of the CoT prompt itself.

In Experiment 2, we changed the standard prompt to a more open-ended format, asking: "Which one of these would you use to accomplish the task?" This resulted in the models giving longer answers. The CoT prompt remained largely the same, asking models to "Evaluate each option separately before specifying your choice." Rather than testing the models at different temperatures, we set the temperature of all models to 0. This is because Experiment 1 showed that model performance at temperature of 0 was higher and because it makes responses less variable and our results more reproducible. We only collected CoT data in the Standard and Image conditions, as collecting CoT data in the Distractor condition would entail the models separately evaluating 9 options, yielding excessively long responses.

---

[1] For more information on representative sampling using Prolific, see: https://researcher-help.prolific.com/en/article/e6555f.

## 5.2 Results

*Humans compared to models* The overall mean accuracies of humans and the two most powerful LLMs, GPT-4o and Claude 3.5, are shown in Figure 6 below (see also Table 3). Consistent with the results from Experiment 1, in the Standard condition, we found that only the performance of GPT-4o with CoT prompting did not differ significantly from that of humans ($t(212) = 0.95$, $p = 1.00$). All other models performed significantly worse than humans (see Appendix B, Table B1). In the Distractor condition, humans performed significantly better than both GPT-4o ($t(152) = 2.52$, $p = .03$) and Claude 3.5 ($t(152) = 4.28$, $p < .001$). In the Image condition, the performance of GPT-4o, with normal and CoT prompting, did not differ significantly from that of humans ($t(152) = 0.70$, $p = 1.00$ and $t(152) = 0.82$, $p = 1.00$ respectively), while Claude 3.5's performance, with normal and CoT prompting, was significantly lower ($t(152) = 13.44$, $p < .001$ and $t(152) = 9.08$, $p < .001$ respectively).

**Figure 6**
*Mean Accuracy of Humans, the GPT-4o, and Claude 3.5 S*

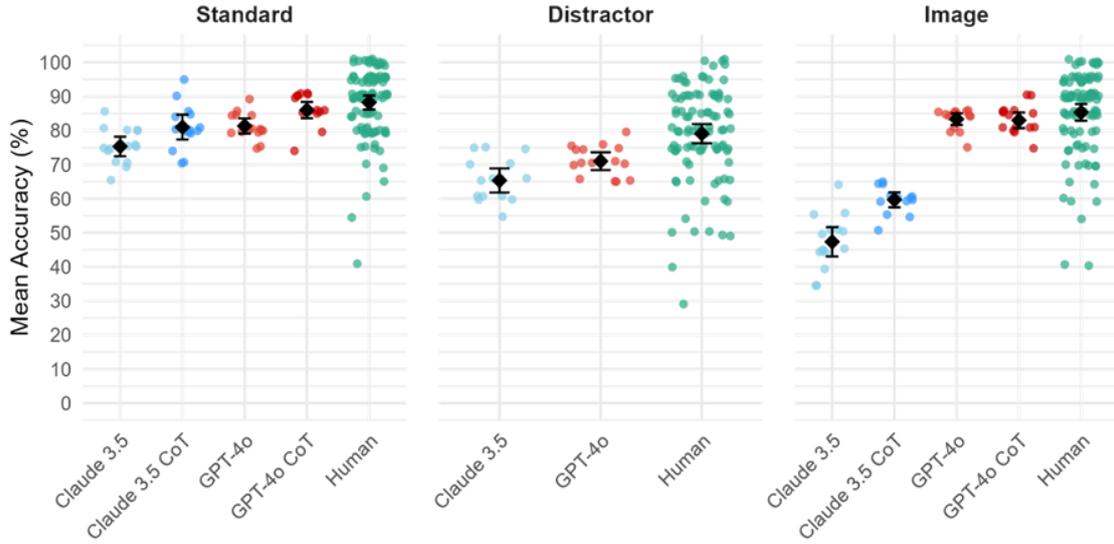

*Note.* Individual data points represent human participant averages and model run averages (i.e., the average performance of a model on a single round of the 20 tasks). Black diamonds represent mean accuracy for each model, with error bars showing the 95% confidence intervals. Individual data points are jittered for better visibility.

**Table 3**
*Overview of mean and sd accuracy (% correct) for humans and all models*

|  | Standard | | Distractor | | Image | |
|---|---|---|---|---|---|---|
| **Model** | *M* | *SD* | *M* | *SD* | *M* | *SD* |
| human | 88.2 | 10.4 | 79.1 | 13.9 | 85.3 | 12.2 |
| GPT-4o | 81.3 | 4.0 | **71.0** | 4.7 | **83.3** | 3.1 |
| GPT-4o CoT | **86.0** | 4.3 | na | na | 83.0 | 4.1 |
| Claude 3.5 | 75.3 | 5.2 | 65.3 | 6.4 | **47.3** | 7.8 |
| Claude 3.5 CoT | 81.0 | 6.6 | na | na | 59.7 | 4.0 |
| GPT-3.5 | **46.7** | 4.5 | 35.3 | 6.1 | na | na |
| GPT-3.5 CoT | 52.3 | 8.2 | na | na | na | na |
| Claude 3 | 51.0 | 8.5 | **32.3** | 4.6 | na | na |
| Claude 3 CoT | 55.0 | 6.6 | na | na | na | na |

*Note.* The highest and lowest LLM score is marked in bold in each condition.

**Effect of Distractor condition** Compared to the four-option Standard (text) condition, the performance of humans and all LLMs declined significantly in the nine-option Distractor condition (see Appendix B, Table B2). Notably, Claude 3's performance decreased by 18.7%, while the other models' performance declined by around 10 and 11% and humans by 9.1%. This significant decline across the board contradicts our expectation that only the earlier models would show this trend.

**Effect of Image condition** The effect of the Image condition was mixed. Compared to the Standard text condition, performance did not decline significantly for humans and GPT-4o in the Image condition (humans: $t(293) = 1.68$, $p = .19$, GPT-4o: $t(70) = -1.34$, $p = .74$). For Claude 3.5, however, performance fell significantly by 28% ($t(70) = 12.53$, $p < .001$). CoT prompting did not affect GPT-4o in the Image condition ($p = 1.00$), while Claude 3.5's performance was significantly strengthened ($p < .001$, for test statistics, see Appendix B, Table B3).

**The effect of CoT prompting** Overall, CoT prompting had a positive effect on performance, increasing the LLMs' scores by 5.35% on average. Only in one out of six cases, for GPT-4o in the image condition, did CoT prompting not improve performance. Statistically, CoT prompting had a significant positive effect in three cases (see Appendix B, Table B3). This suggests that CoT prompting is still effective to some degree, but that the effect is not universal and not always strong. The smaller positive effect of CoT prompting in Experiment 2 may be due to the change in the standard prompt discussed earlier.

**Qualitative analysis of Claude's answers** Here, like we did for the GPT models in Experiment 1, we offer a brief qualitative analysis of some of the answers of Claude 3 and Claude 3.5, both with CoT prompting. By examining some illustrative examples, we hope to give the reader a sense of the ways in which the models succeed and fail at the tasks.

**Figure 7**
*Mean Accuracy of Claude 3 and Claude 3.5 with CoT Prompting Across the 20 Tasks*

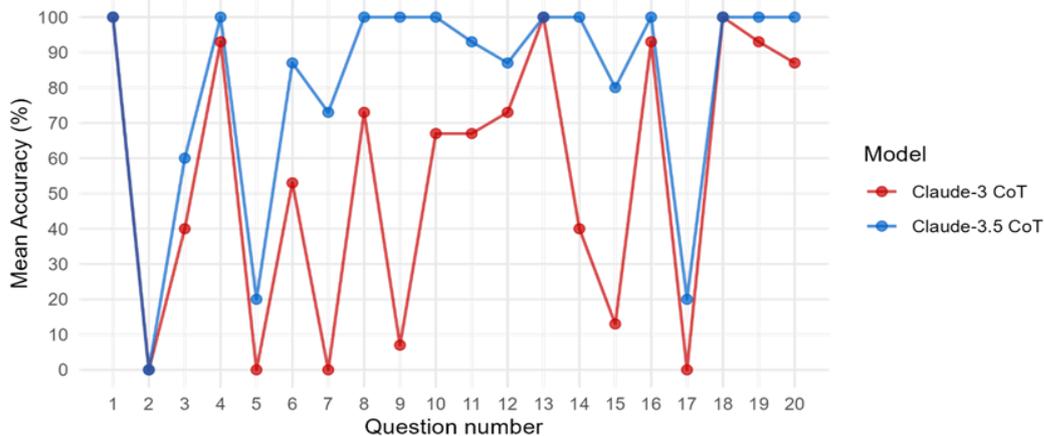

*Question 7 – sweeping the floor* This task involved replacing a broom for sweeping the floor. The afforded object was a wig and the two associated objects were laundry detergent and a bucket. Claude 3 scored 0% on this task and GPT-4o scored 73%. Below are examples of their evaluations of the wig and the bucket:

| Claude 3 | Claude 3.5 |
|---|---|
| "A **wig** is … designed to be worn on the head and would not be a suitable tool for sweeping the floor. The wig's material and construction make it ineffective for gathering and collecting debris." | "A **wig** is made of hair or synthetic fibers and has a somewhat brush-like quality. While not designed for cleaning, its fibers could potentially collect dust and debris if dragged across the floor." |
| "A **wig** is … not suitable for sweeping floors effectively. It may collect some dust and debris, but it would be inefficient and likely leave behind a significant amount of dirt and particles." | "A **wig**: - Pros: It's made of hair-like fibers, which could potentially collect dust and debris. - Cons: It's not designed for cleaning, and using it might damage the wig …. |
| **Bucket:** "While it may not be the most efficient tool for sweeping, it could potentially be used to gather and collect debris from the floor by dragging or pushing it along the surface." | "A **bucket**: - Pros: It has a flat bottom and could potentially be used to push debris into a pile. - Cons: It's not designed for sweeping and would be inefficient. It might scratch the floor." |

Similarly to the difference between GPT-3.5 and GPT-4o, Claude 3.5 showed a stronger appreciation than Claude 3 that the wig is somewhat analogous to a broom's bristles. Claude 3 showed some sensitivity to the analogy, but opted generally for the bucket and chose the irrelevant object, a vase, three times.

***Question 9 – taking a cake out of the oven*** This task required replacing oven mittens to take a hot cake out of the oven. The afforded object was a beach towel, while the two associated objects were a saucepan lid and a chef's hat. Claude 3 scored 7% on this task and Claude 3.5 scored 100%. Below are examples from Claude 3 and Claude 3.5's evaluations of the beach towel and the saucepan lid:

**Claude 3**

"**Towels** are made of thick, absorbent material that can provide some insulation against heat. However, beach towels are not designed for high heat resistance, and there is a risk of the towel catching fire …"

"**Beach towels** are typically thin and may not offer sufficient protection from the high temperature of the oven and the hot cake pan."

"**Saucepan lids** are typically made of metal or heat-resistant materials designed to withstand high temperatures. Using a saucepan lid as a makeshift oven mitt would provide adequate insulation and protection against the heat while allowing you to grip the hot cake securely."

**Claude 3.5**

"A **beach towel**: Pros: Large surface area, thick fabric provides good insulation, can be folded for extra protection. Cons: Might be bulky to handle."

"A **beach towel**: Pros: Large surface area, thick fabric provides insulation Cons: May not be as heat-resistant as specialized oven mitts … You would need to fold the towel several times to create a thick layer between your hands and the hot pan."

"A **saucepan lid**: Pros: Made of heat-resistant material, usually metal or glass. Cons: Small surface area, difficult to grip the cake pan securely, risk of burns."

While Claude 3 recognized that a beach towel could provide some insulation against heat, it only mentioned the possibility of folding the towel once, and it only opted for the beach towel in 1 out of 15 answers. Instead, it went for the associated saucepan lid 14 times and usually claimed that the lid would provide a good "grip" on the cake. Claude 3.5, however, always opted for the beach towel and suggested folding the towel in 14 out of 15 answers.

## 5.3 Discussion

In the Standard condition, the performance of the LLMs relative to humans was similar to the results in the first experiment. Humans scored highest, with GPT-4o close behind and GPT-3.5 far behind. The Claude models also followed this pattern, with Claude 3.5 scoring slightly (but significantly) worse than humans, while Claude 3, Anthropic's earlier model, scored similarly to GPT-3.5, OpenAI's earlier model.

The effect of using images to symbolize the object options instead of words was mixed. Humans and GPT-4o's performance were not significantly affected. This suggests that the sort of reasoning or computation that multimodal LLMs perform to solve the text-condition tasks is not restricted to that modality alone. LLMs can take text and image inputs simultaneously and arrive at the same or similar answers whether it is presented with the word "apple" or an image of an apple. However, it is difficult to say whether these results indicate that the model is able to translate the image input into a text format which it then uses to solve the tasks or if the model reasons using a shared and amodal representational format. There is evidence suggesting that the representations of language and vision transformer models converge to some extent (Li et al., 2024). Moreover, a mechanistic study of extractable features in the Claude 3 Sonnet model found multimodal features, shown by the fact that the same circuits in the model would activate for words and images of the same concept, such as the Golden Gate Bridge (Templeton et al., 2024). This suggests that these LLMs have a shared embedding space for both text and images. Claude 3.5's weak performance in the image condition compared to the text condition suggests that LLMs may differ in their ability to integrate text and image inputs on reasoning tasks.

The effect of including five additional distractor options was significant across the board, for humans and LLMs alike. This contradicted our expectation that only the earlier LLMs would be significantly affected. The performance decline of humans, GPT-4o, and Claude 3.5 was quite similar. This similar sensitivity to the increase in task difficulty suggests that whatever the reasoning process GPT-4o and Claude 3.5 use to solve these types of tasks is somewhat robust. The performance in the distractor condition, even if lower than in the standard condition, might provide somewhat stronger evidence for robust reasoning, given that random guessing would only produce a score of about 11%.

The qualitative review of Claude 3 and Claude 3.5's text output showed similar patterns to those seen for GPT-3.5 and GPT-4o. Claude 3.5's factual and causal knowledge is more detailed, precise, and less error-prone than that of Claude 3. Claude 3.5 was also able to disregard associated and typical options to a much greater degree than Claude 3, suggesting that Claude 3.5 can generalise to a greater extent than its predecessor.

# 6 General Discussion

Our experiments examined the ability of humans and LLMs to deal with unconventional tool use tasks. The experiments were designed to address key issues in the debates about LLMs' abilities. Inspired by Yiu et al.'s (2023) tool innovation task design, we tested LLMs' abilities to reason compositionally and flexibly apply causal knowledge. To better probe the robustness and generality of the LLMs' reasoning, we tested the models in a variety of conditions (different temperatures, prompting techniques, modalities, and levels of difficulty). The task itself involves a variety of objects and situations, from cutting and throwing to baking and gardening. Overall, we found strong progress in LLM responses, whereby GPT-4o and Claude 3.5 performed much better than the earlier GPT-3.5 and Claude 3 models. Notably, GPT-4o performed almost at a human level in the Standard and Image conditions, with Claude 3.5 not far behind in the Standard condition.

Our findings raise many unanswered – and currently unanswerable – questions for the research community. OpenAI and Anthropic, for commercial and safety reasons, do not publish the technical details of their models. This means that we can only speculate about the technical changes behind the noted progress. Given recent trends, it is plausible that scaling has played a major part. Scaling the size of a transformer model, the compute used in training, and the amount of training have all been shown to improve performance on next-token prediction (Kaplan et al., 2020) and on a surprisingly wide variety of tasks. The improved performance we observe in GPT-4o and Claude 3.5 may thus be due to the persisting gains from scaling LLMs. Besides scaling, there is the possibility that "algorithmic improvements", progress in the design of training (Ho et al., 2024) and post-training (Xu et al., 2024) algorithms, has played a part.

**Generalisation and memorisation** Our findings have implications for the issue of memorisation versus generalisation in LLMs. Chang and Bergen (2023), for example, noted in their review of LLMs that it is unclear whether the improvements accompanying larger models are due to better generalisation abilities or the memorisation of a larger number of specific patterns that together improve performance (pp. 30-31). There is evidence that LLMs have memorised some tests and benchmarks, making these invalid tests of reasoning or generalisation (Bordt et al., 2024).

As we discussed earlier, there are three main ways to test generalisation abilities in LLMs: by creating novel versions of familiar tasks, by creating structurally novel tasks, and by creating tasks where there is tension between associative memory and rule-bound reasoning. In this project, we have used the first and third methods of probing generalisation in LLMs. The 20 tasks in this study are novel and they create a tension between associative memory and rule-bound reasoning. However, they are not structurally novel, as the pattern of using familiar objects for new and unfamiliar uses is something the models are likely to have encountered in their training data. The finding that GPT-4o with CoT prompting performed at a human level in the Standard text and Image conditions strongly suggests that current state-of-the-art LLMs can to some extent generalise.

However, one might still argue that Claude 3.5 and GPT-4o's strong performance is due to improved memorisation abilities. Larger models have been found to memorise better (Tirumala et al., 2022), and it is possible that better memorisation is key to GPT-4o's performance. What speaks against this view is our qualitative analysis of the LLMs' responses. When examining their output, we found that the models usually did not struggle to identify causally relevant properties required to solve the tasks. In this sense, their memorised knowledge about the relevant task properties was adequate. GPT-3.5 and Claude 3's problem, relative to GPT-4o and Claude 3.5, was that they struggled to decompose the afforded object into abstract properties that it shared with the typical tool, and they struggled to disregard common use patterns. This does not deny that better and more precise memory helped the later models outperform their predecessors, but it suggests that the ability to disregard memorised patterns and to reason abstractly and compositionally also played an important role in solving the tasks.

This is consistent with Wu et al.'s (2023) findings that while GPT-4 performs better on familiar than unfamiliar task structures, it still performs significantly above chance on many novel tasks. However, the finding that LLMs can to some degree generalise on these tasks does not imply universal generalisation abilities. There are still many tasks on which LLMs struggle (e.g., Philip & Hemang, 2024).

**Characterising abilities** Our findings and argument that LLMs use compositional reasoning align with nascent research on LLMs' internal representations. According to Pavlick (2023), compositionality is dependent upon the existence of modular representations of concepts and their roles. While research on the internal representations of LLMs is still in its beginnings, multiple studies have already identified parameters (parts of an LLM's artificial neural network) that correspond to concepts such as specific animals, time, abstract "part-of" relations, and functions (Pavlick, 2023; Geva et al., 2021; Todd et al., 2024). Moreover, researchers have been able to manipulate LLM output in targeted ways by intervening on these internal parameters, showing that the identified representations have a causal impact on the models' behaviour (Geiger et al., 2021). Given the evidence for the existence of abstract representations in LLMs, it is plausible that LLMs' generalisation and innovative skills stem, at least in part, from the ability to flexibly recombine these abstract representations; in other words, from the ability to reason compositionally.

The models' performance and output also showed that they have encoded enormous amounts of causal knowledge. This knowledge was significantly more accurate in GPT-4o and Claude 3.5 than in GPT-3.5 and Claude 3. To a large extent, the causal knowledge relevant for the tasks concerned the properties and affordances of various materials and shapes. Our results align with previous findings that LLMs perform increasingly well on causal reasoning and affordance tasks (Kiciman et al., 2023; Jones et al., 2022).

Does this suggest that LLMs have human-like causal knowledge? We argue that the results of our behavioural study cannot address this question; it requires further mechanistic research (Rai et al., 2024). However, we note that it is important to separate knowledge from intelligence itself, the process that *generates* the knowledge (Chollet, 2019). While LLMs have encoded more causal knowledge than a single human ever could, this is not the same thing as the ability to generate causal models and knowledge. Moreover, the nature of LLMs' knowledge and how coherent their models remain unclear. One mechanistic study by Vafa et al. (2024), for example, found that generative models trained to navigate a city may achieve very strong performance without forming accurate models of the city. Another study investigated the algorithms used by LLMs to solve arithmetic tasks and found "neither robust algorithms nor memorization", but instead a "bag of heuristics" approach (Nikankin et al., 2024). Thus, while LLMs show through their output that they have encoded human-generated causal knowledge, the mechanistic nature of this knowledge and how it relates to different prompting strategies (Wang et al., 2023) remain areas for further investigation.

**Limitations and future research** Our behavioural studies, while revealing performance patterns, cannot definitively establish the mechanisms enabling LLMs' success on these tasks. While emerging mechanistic research suggests LLMs develop modular concept representations (Pavlick, 2023; Todd et al., 2024), the nature of their causal knowledge and reasoning processes requires further investigation. Additionally, given LLMs' vast training data (Liu et al., 2024), ensuring complete task novelty remains challenging, highlighting the need for methods to verify task originality before experimental design (Bordt et al., 2024). Another limitation lies in the text-based nature of our evaluation. As Rutar et al. (2024) highlight, affordances are about the ability to interact with objects and learn from these interactions, not just the ability to generate accurate text about objects. Although our findings demonstrate that LLMs can make sophisticated judgments about object affordances, this is not the same as learning and applying a causal model in the real world. Further research is necessary to evaluate how well multimodal LLMs can manipulate and use objects in real or simulated environments.

# 7 Conclusion

As the progress in AI and LLMs continues at a rapid pace, understanding the capabilities and limitations of these systems becomes increasingly important. We have investigated the compositional and causal reasoning abilities in the domain of object affordances using a tool innovation task. Our findings suggest that LLMs have progressed significantly in their ability to solve these tasks, and that models like GPT-4o and Claude 3.5 perform at a human and near-human level in certain experimental conditions. This suggests that models are able, at least to a degree, to reason compositionally and apply causal knowledge flexibly in ways that generalise to new situations. As the LLM debate moves forward, further mechanistic research will be needed to clarify how LLMs represent their knowledge and solve reasoning tasks.

## Appendix A

**Table A1**
Overview of the 20 tool innovation questions used in the experiments. Options only present in the Distractor condition are *italicized*.

| Q | goal | typical tool | afforded object | associated objects | irrelevant objects |
|---|---|---|---|---|---|
| 1 | keep your body warm | jacket | curtains | umbrella<br>boxer shorts<br>*baseball cap*<br>*leather gloves* | bowl<br>*charger*<br>*crayon*<br>*letter opener* |
| 2 | dig a hole | spade | table tennis racket | rake<br>garden hose<br>*wheelbarrow*<br>*lawnmower* | a drying rack<br>*notebook*<br>*lighter*<br>*eraser* |
| 3 | hammer a nail | hammer | saucepan | screwdriver<br>saw<br>*meter stick*<br>*chisel* | picture frame<br>*bicycle wheel*<br>*jug*<br>*nail clipper* |
| 4 | transport water | glass | swimming cap | plate<br>fork<br>*flour sifter*<br>*napkin* | rubber bands<br>*hairbrush*<br>*flashlight*<br>*shoehorn* |
| 5 | cut a cucumber into cubes | kitchen knife | ruler | peeler<br>whisk<br>*mortar and pestle*<br>*potato masher* | ash tray<br>*dustpan*<br>*plastic bag*<br>*fly swatter* |
| 6 | rest your head | small pillow | kitchen roll | sleep mask<br>bedside table<br>*sleeping pill*<br>*pillowcase* | book<br>*compass*<br>*corkscrew*<br>*crutch* |
| 7 | sweep the floor | broom | wig | bucket<br>laundry detergent<br>*broom holder*<br>*spray bottle* | vase<br>*baking tray*<br>*envelope*<br>*salt shaker* |
| 8 | draw a straight line | ruler | DVD case | paperclips<br>pencil sharpener<br>*staple remover*<br>*correction tape* | plant pot<br>*razor*<br>*pepper mill*<br>*shoe* |
| 9 | take a baked and hot cake out of the oven | oven mittens | beach towel | chef's hat<br>saucepan lid<br>*measuring cup*<br>*cheese knife* | balloon<br>*paintbrush*<br>*needle*<br>*candle* |
| 10 | roll out a pizza dough | rolling pin | baseball bat | pizza cutter<br>measuring spoon<br>*pastry brush*<br>*garlic press* | backpack<br>*crowbar*<br>*comb*<br>*jewellery box* |
| 11 | fasten gift wrapping paper around a gift | scotch tape | fishing line | scissors<br>rubber stamp<br>*notepad*<br>*quill pen* | chopsticks<br>*tablecloth*<br>*whiteboard sponge*<br>*bicycle pump* |
| 12 | write down a short note | post-it note | coffee filter | ink cartridge<br>glue stick<br>*hole puncher*<br>*stapler* | toothbrush<br>*helmet*<br>*pair of skis*<br>*sunglasses* |
| 13 | serve fruit on a plate | serving plate | frisbee | apple slicer<br>juice carton<br>*blender*<br>*ice cream scoop* | duffel bag<br>*map*<br>*scythe*<br>*teddy bear* |
| 14 | fasten a piece of paper to a wall | Blu Tack | plasters | folder<br>rubber bands<br>*whiteboard*<br>*pen* | boots<br>*water bottle*<br>*wrench*<br>*bar soap* |
| 15 | practice doing baseball tosses with a friend | baseball | apple | basketball<br>bowling ball<br>*table tennis ball*<br>*volleyball* | broom<br>*blanket*<br>*bed sheet*<br>*thermos* |
| 16 | keep your ears warm | beanie | audio headset | skiing glasses<br>winter gloves<br>*socks*<br>*fur boots* | keychain<br>*padlock*<br>*grater*<br>*tweezers* |
| 17 | replace a missing king piece on a chess board | the king piece | one pound coin | crown<br>sceptre<br>*orb*<br>*throne* | coffee cup<br>*CD disk*<br>*brick*<br>*ladder* |
| 18 | beat the dust out of a rug | rug beater | tennis racket | feather duster<br>sponge<br>*stain remover*<br>*clothespin* | matchbox<br>*button*<br>*mirror*<br>*hand watch* |
| 19 | mark where you are in a book | bookmark | shoelace | book stand<br>highlighter<br>*library catalogue*<br>*reading lamp* | chopping board<br>*belt*<br>*pushpin*<br>*magnifying glass* |
| 20 | carry laundry from one place to another | laundry basket | picnic blanket | toiletry bag<br>hanger<br>*iron*<br>*fabric softener* | can opener<br>*lunchbox*<br>*wallet*<br>*toothpick* |

# Appendix B

**Table B1**
*Pairwise Comparisons of Mean Accuracy Between Humans and the LLMs in the Standard condition*

| Contrast | Estimate | SE | df | t | p |
| --- | --- | --- | --- | --- | --- |
| GPT-4o CoT vs. Human | -0.02 | 0.02 | 212 | 0.95 | 1.00 |
| GPT-4o vs. Human | -0.07 | 0.02 | 212 | 2.94 | .03 |
| GPT-3.5 CoT – Human | -0.36 | 0.02 | 212 | 15.32 | < .001 |
| GPT-3.5 vs. Human | -0.42 | 0.02 | 212 | 17.74 | < .001 |
| Claude 3.5 CoT vs. Human | -0.07 | 0.02 | 212 | 3.08 | .02 |
| Claude 3.5 vs. Human | -0.13 | 0.02 | 212 | 5.50 | < .001 |
| Claude 3 CoT vs. Human | -0.33 | 0.02 | 212 | 14.19 | < .001 |
| Claude 3 vs. Human | -0.37 | 0.02 | 212 | 15.89 | < .001 |

*Note.* Estimate, i.e. the difference between the means, is also here expressed as the ratio of correct responses.

**Table B2**
*Pairwise Comparisons of Mean Accuracy in Standard and Distractor conditions*

| Contrast (standard vs. distractor) | Estimate | SE | df | t | p |
| --- | --- | --- | --- | --- | --- |
| Humans | -0.09 | 0.02 | 293 | 5.25 | < .001 |
| GPT-4o | -0.10 | 0.01 | 70 | 6.93 | < .001 |
| GPT-3.5 | -0.11 | 0.02 | 42 | 4.81 | < .001 |
| Claude 3.5 | -0.10 | 0.02 | 70 | 4.47 | < .001 |
| Claude 3 | -0.19 | 0.02 | 42 | 7.59 | < .001 |

*Note.* Estimate, i.e. the difference between the means, is also here expressed as the ratio of correct responses.

**Table B3**
*Pairwise Comparisons of Mean Accuracy with and without CoT Prompting*

| Contrast (CoT vs. normal prompt) | Estimate | SE | df | t | p |
| --- | --- | --- | --- | --- | --- |
| GPT-4o Standard | 0.05 | 0.01 | 70 | 3.13 | .01 |
| GPT-4o Image | -0.00 | 0.01 | 70 | 0.22 | 1.00 |
| GPT-3.5 Standard | 0.06 | 0.01 | 42 | 2.40 | .04 |
| Claude 3.5 Standard | 0.06 | 0.02 | 70 | 2.54 | .054 |
| Claude 3.5 Image | 0.12 | 0.02 | 70 | 5.52 | < .001 |
| Claude 3 Standard | 0.04 | 0.02 | 42 | 1.63 | .22 |

*Note.* Estimate, i.e. the difference between the means, is also here expressed as the ratio of correct responses.